\def\year{2020}\relax
\documentclass[letterpaper]{article} % DO NOT CHANGE THIS
\usepackage{aaai20}  % DO NOT CHANGE THIS
\usepackage{times}  % DO NOT CHANGE THIS
\usepackage{helvet} % DO NOT CHANGE THIS
\usepackage{courier}  % DO NOT CHANGE THIS
\usepackage[hyphens]{url}  % DO NOT CHANGE THIS
\usepackage{graphicx} % DO NOT CHANGE THIS
\usepackage{graphicx}  % DO NOT CHANGE THIS
\title{Modelling Semantic Categories using Conceptual Neighborhood}
\author{Zied Bouraoui\\
CRIL - U.\ Artois - CNRS\\
zied.bouraoui@cril.fr\\
\And
Jose Camacho-Collados\\
Cardiff University, UK\\
camachocolladosj@cardiff.ac.uk
\And
Luis Espinosa-Anke\\
Cardiff University, UK\\
espinosa-ankel@cardiff.ac.uk
\And
Steven Schockaert\\
Cardiff University, UK\\
schockaerts1@cardiff.ac.uk
}

\usepackage{latexsym}
\usepackage{graphicx}
\usepackage{amsmath,amssymb}
\usepackage{booktabs}
\usepackage{longtable}
\usepackage{array}
\usepackage{multirow}
\usepackage{wrapfig}
\usepackage{float}
\usepackage{colortbl}
\usepackage{pdflscape}
\usepackage{tabu}
\usepackage{threeparttable}
\usepackage{threeparttablex}
\usepackage[normalem]{ulem}
\usepackage{makecell}
\usepackage{graphicx}
\usepackage{xcolor}
\usepackage{rotating}
\usepackage{url}
\usepackage{subcaption}
\usepackage{tabularx}
\usepackage[export]{adjustbox}
\usepackage{array}

\newcommand{\citet}[1]{\citeauthor{#1}~\shortcite{#1}}
\newcommand{\newcite}[1]{\citeauthor{#1}~\shortcite{#1}}

\usepackage{graphicx}

\usepackage{url}
 
\begin{document}

\maketitle

\begin{abstract}
While many methods for learning vector space embeddings have been proposed in the field of Natural Language Processing, these methods typically do not distinguish between categories and individuals. Intuitively, if individuals are represented as vectors, we can think of categories as (soft) regions in the embedding space. Unfortunately, meaningful regions can be difficult to estimate, especially since we often have few examples of individuals that belong to a given category. To address this issue, we rely on the fact that different categories are often highly interdependent. In particular, categories often have conceptual neighbors, which are disjoint from but closely related to the given category (e.g.\ fruit and vegetable). Our hypothesis is that more accurate category representations can be learned by relying on the assumption that the regions representing such conceptual neighbors should be adjacent in the embedding space. We propose a simple method for identifying conceptual neighbors and then show that incorporating these conceptual neighbors indeed leads to more accurate region based representations.
\end{abstract}

\section{Introduction}
Vector space embeddings are commonly used to represent entities in fields such as machine learning (ML) \cite{NIPS20135071}, natural language processing (NLP) \cite{camacho2016nasari}, information retrieval (IR) \cite{ASI:ASI1} and cognitive science \cite{Gardenfors:conceptualSpaces}. An important point, however, is that such representations usually represent both individuals and categories as vectors \cite{ma2016label,zheng2016joint,boleda2017instances}. Note that in this paper, we use the term \textit{category} to denote natural groupings of individuals, as it is used in cognitive science, with \textit{individuals} referring to the objects from the considered domain of discourse. For example, the individuals \textit{carrot} and \textit{cucumber} belong to the \textit{vegetable} category\footnote{Note that the same entity could be treated as an individual or a category depending on the context; e.g.\ \textit{carrot} is a category of physical objects, but an instance of the \textit{vegetable} category.}. We use the term \textit{entities} as an umbrella term covering both individuals and categories.
%
%Most existing approaches are essentially aimed at modelling similarity. In NLP, for instance, word embeddings are usually learned from co-occurrence statistics (i.e.\ words are represented as similar vectors if they tend to co-occur with the same words), while in psychological studies semantic spaces are often directly learned from human similarity judgements. 

Given that a category corresponds to a set of individuals (i.e.\ its instances), modelling them as (possibly imprecise) regions in the embedding space seems more natural than using vectors. In fact, it has been shown that the vector representations of individuals that belong to the same category are indeed often clustered together in learned vector space embeddings \cite{gupta2015distributional,DBLP:conf/sigir/JameelBS17}. 
%\red{Perhaps a small definition of "category", which can be ambiguous, with an example could help at this point.} 
The view of categories being regions is also common in cognitive science \cite{Gardenfors:conceptualSpaces}. However, learning region representations of categories is a challenging problem, because we typically only have a handful of examples of individuals that belong to a given category. One common assumption is that natural categories can be modelled using \textit{convex} regions \cite{Gardenfors:conceptualSpaces}, which simplifies the estimation problem. For instance, based on this assumption, \citet{DBLP:conf/aaai/BouraouiJS17} modelled categories using Gaussian distributions and showed that these distributions can be used for knowledge base completion. Unfortunately, this strategy still requires a relatively high number of training examples to be successful. 

However, when learning categories, humans do not only rely on examples. For instance, there is evidence that when learning the meaning of nouns, children rely on the default assumption that these nouns denote mutually exclusive categories \cite{markman1990constraints}. In this paper, we will in particular take advantage of the fact that many natural categories are organized into so-called \emph{contrast sets} \cite{goldstone1996isolated}. These are sets of closely related categories which exhaustively cover some sub-domain, and which are assumed to be mutually exclusive; e.g.\ the set of all common color names, the set $\{\text{fruit},\text{vegetable}\}$ or the set $\{\text{NLP}, \text{IR}, \text{ML}\}$. Categories from the same contrast set often compete for coverage. For instance, we can think of the NLP domain as consisting of research topics that involve processing textual information \emph{which are not covered by the IR and ML domains}. Categories which compete for coverage in this way are known as \emph{conceptual neighbors} \cite{Freksa:1991}; e.g.\ NLP and IR, red and orange, fruit and vegetable. Note that the exact boundary between two conceptual neighbors may be vague (e.g.\ tomato can be classified as fruit or as vegetable).

\begin{figure}
\centering
\includegraphics[width=150pt]{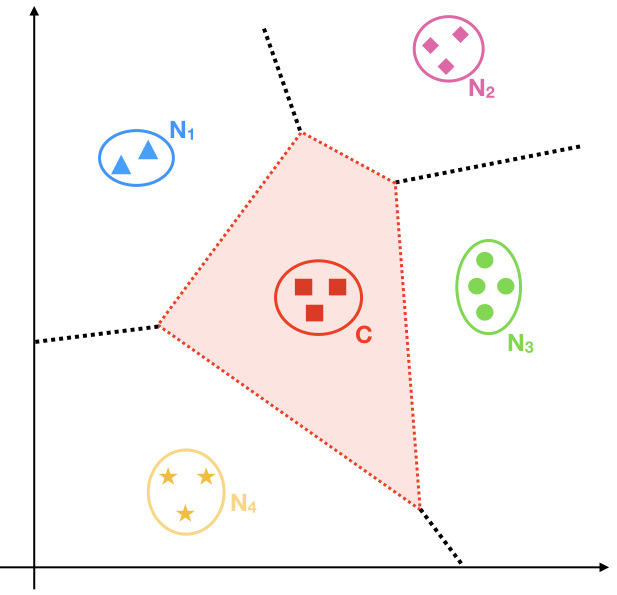}
\caption{Using conceptual neighborhood for estimating category boundaries. \label{figDiagram}}
\end{figure}

In this paper, we propose a method for learning region representations of categories which takes advantage of conceptual neighborhood, especially in scenarios where the number of available training examples is small. The main idea is illustrated in Figure \ref{figDiagram}, which depicts a situation where we are given some examples of a target category $C$ as well as some related categories $N_1,N_2,N_3,N_4$. If we have to estimate a region from the examples of $C$ alone, the small elliptical region shown in red would be a reasonable choice. More generally, a standard approach would be to estimate a Gaussian distribution from the given examples. However, vector space embeddings typically have hundreds of dimensions, while the number of known examples of the target category is often far lower (e.g.\ 2 or 3). In such settings we will almost inevitably underestimate the coverage of the category\footnote{Note that $k$ examples span a subspace of at most $k-1$ dimensions, and can thus not provide us with any information about the variance along directions which are orthogonal to that subspace.}. However, in the example from Figure \ref{figDiagram}, if we take into account the knowledge that $N_1,N_2,N_3,N_4$ are conceptual neighbors of $C$, the much larger, shaded region becomes a more natural choice for representing $C$. Indeed, the fact that e.g.\ $C$ and $N_1$ are conceptual neighbors suggests that any point in between the examples of these categories needs to be contained either in the region representing $C$ or the region representing $N_1$. In the spirit of prototype approaches to categorization \cite{rosch1973natural}, without any further information it makes sense to assume that their boundary is more or less half-way in between the known examples.

The contribution of this paper is two-fold. First, we propose a method for identifying conceptual neighbors from text corpora. We essentially treat this problem as a standard text classification problem, by relying on categories with large numbers of training examples to generate a suitable distant supervision signal. Second, we show that the predicted conceptual neighbors can effectively be used to learn better category representations. 

%Let us assume that the categories are organised in a taxonomy, and let $C_1,...,C_n$ be the set of categories which share some parent $P$ in that taxonomy. In many cases, we can then think of $C_1,...,C_n$ as defining a contrast set. To learn a representation of such a contrast set $C_1,...,C_n$, we could refine the representation of $P$ by training a discriminative model on the available examples of the categories $C_1,...,C_n$. For instance, in line with the prototype theory model of categorization \cite{rosch1973natural}, many psychological studies assume that we only have a single prototypical example for each category and then compute the Voronoi tessellation induced by these examples to obtain region boundaries; see \cite{Gardenfors:conceptualSpaces} for an overview.

%For a given contrast set $C_1,...,C_n$, we may thus learn meaningful region boundaries even if we only have very few examples of each category. However, 0

%******************************
\section{Related Work}
In distributional semantics, categories are frequently modelled as vectors. For example, \citet{DBLP:journals/corr/abs-1808-01662} study the problem of deciding for a word pair $(i,c)$ whether $i$ denotes an instance of the category $c$, which they refer to as \emph{instantiation}. They treat this problem as a binary classification problem, where e.g.\ the pair (AAAI, conference) would be a positive example, while (conference, AAAI) and (New York, conference) would be negative examples. Different from our setting, their aim is thus essentially to model the instantiation relation itself, similar in spirit to how hypernymy has been modelled in NLP \cite{weeds2014learning,roller2014inclusive}. To predict instantiation, they use a simple neural network model which takes as input the word vectors of the input pair $(i,c)$. They also experiment with an approach that instead models a given category as the average of the word vectors of its known instances and found that this led to better results. 

A few authors have already considered the problem of learning region representations of categories. Most closely related, \citet{DBLP:conf/ijcai/BouraouiS18} model ontology concepts using Gaussian distributions. %, which they estimate from the vector representations of their known instances. 
In \newcite{DBLP:conf/ecai/JameelS16}, a model is presented which embeds Wikipedia entities such that entities which have the same WikiData type are characterized by some region within a low-dimensional subspace of the embedding. Within the context of knowledge graph embedding, several approaches have been proposed that essentially model semantic types as regions \cite{DBLP:conf/naacl/NeelakantanC15,DBLP:conf/acl/GuoWWWG15}. A few approaches have also been proposed for modelling word meaning using regions \cite{Erk:2009:RWR:1596374.1596387,DBLP:conf/conll/JameelS17} or Gaussian distributions \cite{DBLP:journals/corr/VilnisM14}. Along similar lines, several authors have proposed approaches inspired by probabilistic topic modelling, which model latent topics using Gaussians \cite{das2015gaussian} or related distributions \cite{batmanghelich2016nonparametric}.

On the other hand, the notion of conceptual neighborhood has been covered in most detail in the field of spatial cognition, starting with the influential work of \citet{Freksa:1991}. In computational linguistics, moreover, this representation framework aligns with lexical semantics traditions where word meaning is constructed in terms of \textit{semantic decomposition}, i.e.\ lexical items being minimally decomposed into structured forms (or templates) rather than sets of features \cite{pustejovsky1991generative}, effectively mimicking a sort of conceptual neighbourhood. In Pustejovsky's \textit{generative lexicon}, a set of ``semantic devices'' is proposed such that they behave in semantics similarly as grammars do in syntax. Specifically, this framework considers the \textit{qualia} structure of a lexical unit as a set of expressive semantic distinctions, the most relevant for our purposes being the so-called \textit{formal role}, which is defined as ``that which distinguishes the object within a larger domain'', e.g.\ shape or color. This semantic interplay between cognitive science and computational linguistics gave way to the term \textit{lexical coherence}, which has been used for contextualizing the meaning of words in terms of how they relate to their conceptual neighbors \cite{wellner2006classification}, or by providing expressive lexical semantic resources in the form of ontologies \cite{pustejovsky2006towards}.

%\footnote{The terminology between cognition and linguistics, as expected, sometimes differs. In lexical semantics and phonetics, for instance, we find the term \textit{lexical neighborhood} being used instead of \textit{conceptual neighborhood}.}

%******************************
\section{Model Description}
\label{model}

Our aim is to introduce a model for learning region-based category representations which can take advantage of knowledge about the conceptual neighborhood of that category. Throughout the paper, we focus in particular on modelling categories from the BabelNet taxonomy \cite{navigli2012babelnet}, although the proposed method can be applied to any resource which (i) organizes categories in a taxonomy and (ii) provides examples of individuals that belong to these categories. Selecting BabelNet as our use case is a natural choice, however, given its large scale and the fact that it integrates many lexical and ontological resources. 

As the possible conceptual neighbors of a given BabelNet category $C$, we consider all its siblings in the taxonomy, i.e.\ all categories $C_1,...,C_k$ which share a direct parent with $C$. To select which of these siblings are most likely to be conceptual neighbors, we look at mentions of these categories in a text corpus. As an illustrative example, consider the pair (hamlet,village) and the following sentence\footnote{\url{https://en.wikipedia.org/wiki/Hamlet_(place)}}:
\begin{quote}
\emph{In British geography, a \underline{hamlet} is considered smaller than a \underline{village} and ...}
\end{quote}
From this sentence, we can derive that \emph{hamlet} and \emph{village} are disjoint but closely related categories, thus suggesting that they are conceptual neighbors. However, training a classifier that can identify conceptual neighbors from such sentences is complicated by the fact that conceptual neighborhood is not covered in any existing lexical resource, to the best of our knowledge, which means that large sets of training examples are not readily available. 
To address this lack of training data, we rely on a distant supervision strategy. The central insight is that for categories with a large number of known instances, we can use the embeddings of these instances to check whether two categories are conceptual neighbors. In particular, our approach involves the following three steps:
\begin{enumerate}
\item Identify pairs of categories that are likely to be conceptual neighbors, based on the vector representations of their known instances. 
\item Use the pairs from Step 1 to train a classifier that can recognize sentences which indicate that two categories are conceptual neighbors.
\item Use the classifier from Step 2 to predict which pairs of BabelNet categories are conceptual neighbors and use these predictions to learn category representations.
\end{enumerate}
Note that in Step 1 we can only consider BabelNet categories with a large number of instances, while the end result in Step 3 is that we can predict conceptual neighborhood for categories with only few known instances. We now discuss the three aforementioned steps one by one.

% Note that noisy labels not so problematic as unrelated words often not in the same sentence

\subsection{Step 1: Predicting Conceptual Neighborhood from Embeddings}\label{secDistantSupervision}

Our aim here is to generate distant supervision labels for pairs of categories, indicating whether they are likely to be conceptual neighbors. These labels will then be used in Section \ref{secTextPrediction} to train a classifier for predicting conceptual neighborhood from text. 

Let $A$ and $B$ be siblings in the BabelNet taxonomy. If enough examples of individuals belonging to these categories are provided in BabelNet, we can use these instances to estimate high-quality representations of $A$ and $B$, and thus estimate whether they are likely to be conceptual neighbors. In particular, we split the known instances of $A$ into a training set $I^A_{\textit{train}}$  and test set $I^A_{\textit{test}}$, and similar for $B$. We then train two types of classifiers. The first classifier estimates a Gaussian distribution for each category, using the training instances in $I^A_{\textit{train}}$ and $I^B_{\textit{train}}$ respectively. This should provide us with a reasonable representation of $A$ and $B$ regardless of whether they are conceptual neighbors. In the second approach, we first learn a Gaussian distribution from the joint set of training examples $I^A_{\textit{train}} \cup I^B_{\textit{train}}$ and then train a logistic regression classifier to separate instances from $A$ and $B$. In particular, note that in this way, we directly impose the requirement that the regions modelling $A$ and $B$ are adjacent in the embedding space (intuitively corresponding to two halves of a Gaussian distribution). We can thus expect that the second approach should lead to better predictions than the first approach if $A$ and $B$ are conceptual neighbors and to worse predictions if they are not. In particular, we propose to use the relative performance of the two classifiers as the required distant supervision signal for predicting conceptual neighborhood.

We now describe the two classification models in more detail, after which we explain how these models are used to generate the distant supervision labels.

\begin{enumerate}

\item{\textbf{Gaussian Classifier}}\label{secGaussian}
The first classifier follows the basic approach from \citet{DBLP:conf/ijcai/BouraouiS18}, where Gaussian distributions were similarly used to model WikiData categories. In particular, we estimate the probability that an individual $e$ with vector representation $\mathbf{e}$ is an instance of the category $A$ as follows:
\begin{align*}
P(A|\mathbf{e}) &= \lambda_A \cdot \frac{f(\mathbf{e} | A)}{f(\mathbf{e})}
\end{align*}
\noindent where $\lambda_A$ is the prior probability of belonging to category $A$, the likelihood $f(\mathbf{e} | A)$ is modelled as a Gaussian distribution and $f(\mathbf{e})$ will also be modelled as a Gaussian distribution. Intuitively, we think of the Gaussian $f(. | A)$ as defining a soft region, modelling the category $A$. Given the high-dimensional nature of typical vector space embeddings, we use a mean field approximation:
\begin{align*}
f(\mathbf{e} | A)&= \prod_{i=1}^{d} f_i(e_i | A)
\end{align*}
Where $d$ is the number of dimensions in the vector space embedding, $e_i$ is the $i^{\textit{th}}$ coordinate of $\mathbf{e}$, and  $f_i(. | A)$ is a univariate Gaussian. To estimate the parameters $\mu_i$ and $\sigma_i^2$ of this Gaussian, we use a Bayesian approach with a flat prior:
\begin{align*}
f_i(e_i | A) &=
\int G(e_i;\mu_i,\sigma_i^2) \textit{NI$\chi^{2}$}(\mu,\sigma^2) d\mu d\sigma
\end{align*}
where $G(e_i;\mu_i,\sigma_i^2)$ represents the Gaussian distribution with mean $\mu_i$ and variance $\sigma_i^2$ and \textit{NI$\chi^{2}$} is the normal inverse-$\chi^{2}$ distribution. In other words, instead of using a single estimate of the mean $\mu$ and variance $\sigma_2$ we average over all plausible choices of these parameters. The use of the normal inverse-$\chi^{2}$ distribution for the prior on $\mu_i$ and $\sigma_i^2$ is a common choice, which has the advantage that the above integral simplifies to a Student-t distribution. In particular, we have:
\begin{align*}
f_i(e_i | A) &= t_{n-1}\left(\overline{x_i},\frac{(n+1)\sum_{j=1}^n (a_i^{j}-\overline{x_i})^2}{n(n-1)}\right)
\end{align*}
where we assume $I^A_{\textit{train}}= \{a_1,...,a_n\}$, $a_i^j$ denotes the $i^{\textit{th}}$ coordinate of the vector embedding of $a_j$, $\overline{x_i} = \frac{1}{n}\sum_{j=1}^n a_i^j$ and $t_{n-1}$ is the Student t-distribution with $n-1$ degrees of freedom. The probability $f(\mathbf{e})$ is estimated in a similar way, but using all BabelNet instances. The prior $\lambda_A$ is tuned based on a validation set. 
%, let $I$ be the available instances of the parent of $A$. We learn $\lambda_A$ by maximizing the following log-likelihood:
%$$
%\sum_{e\in I^A_{\textit{train}}} \log P(A| \mathbf{e}) + \sum_{e\in I {\setminus} I^A_{\textit{train}}} \log(1{-} P(A | \mathbf{e}))
%$$
%However, for efficiency, rather than considering all elements from $I {\setminus} I^A_{\textit{train}}$ in the second summation, we select a random subset of size $3\cdot |I^A_{\textit{train}}|$.
Finally, we classify $e$ as a positive example if $P(A|\mathbf{e}) > 0.5$.

\item {\textbf{GLR Classifier.}}\label{secGLR}
We first train a Gaussian classifier as in Section \ref{secGaussian}, but now using the training instances of both $A$ and $B$. 
Let us denote the probability predicted by this classifier as $P(A\cup B | \textbf{e})$. The intuition is that entities for which this probability is high should either be instances of $A$ or of $B$, provided that $A$ and $B$ are conceptual neighbors. If, on the other hand, $A$ and $B$ are not conceptual neighbors, relying on this assumption is likely to lead to errors (i.e.\ there may be individuals whose representation is in between $A$ and $B$ which are not instances of either), which is what we need for generating the distant supervision labels. If $P(A\cup B | \textbf{e}) > 0.5$, we assume that $e$ either belongs to $A$ or to $B$. To distinguish between these two cases, we train a logistic regression classifier, using the instances from $I^A_{\textit{train}}$ as positive examples and those from $I^B_{\textit{train}}$ as negative examples. 
Putting everything together, we thus classify $e$ as a positive example for $A$ if $P(A\cup B | \textbf{e})>0.5$ and $e$ is classified as a positive example by the logistic regression classifier. Similarly, we classfiy $e$ as a positive example for $B$ if $P(A\cup B | \textbf{e})>0.5$ and $e$ is classified as a negative example by the logistic regression classifier. We will refer to this classification model as GLR (Gaussian Logistic Regression).

\end{enumerate}

%and an instance $s$, we first use Gaussian classifier to estimate how likely the instance $s$ belongs to the concept $S_i$ or $S_j$. The Gaussian classifier is learned as described above on the instances of $S_i$ and $S_j$ ($S_i \cup S_j$). Let $P(S_{ij}|v_s)$ be the probability that $s$ is an instance of $S_i$ or $S_j$.  If $P(S_{ij}|v_s) > \tau_{S_{ij}}$, then to decide whether $s$ is a valid instance of $S_i$ or $S_j$, we learn a logistic regression classifier on the instance of $S_i$ and $S_j$. This classifier permits to discriminates between the instances of the two siblings.  

\subsubsection{Generating Distant Supervision Labels}\label{secCombiningPredictions}
%\todo{Finally, Given a concept $S$, to decide whether a given instance belongs to it or not, we tune a threshold $\tau_S$ on the probability of an instance being valid. 
%Let $(S_i,S_j)$ be two sibling concepts and $v_s$ a vector representation of an instance $s$. If $P(S_i|v_s) < \tau_{S_i}$ and $P(S_j|v_s) < \tau_{S_j}$ then $s$ is not an instance of these concepts, otherwise $s$ belongs to the concept where it has the highest probability (an above the threshold).}

To generate the distant supervision labels, we consider a ternary classification problem for each pair of siblings $A$ and $B$. In particular, the task is to decide for a given individual $e$ whether it is an instance of $A$, an instance of $B$, or an instance of neither (where only disjoint pairs $A$ and $B$ are considered). For the Gaussian classifier, we predict $A$ iff $P(A|\mathbf{e})>0.5$  and $P(A|\mathbf{e}) > P(B|\mathbf{e})$. For the GLR classifier, we predict $A$ if $P(A\cup B|\mathbf{e}) >0.5$  and the associated logistic regression classifier predicts $A$. The condition for predicting $B$ is analogous. The test examples for this ternary classification problem consist of the elements from $I^A_{\textit{test}}$ and $I^B_{\textit{test}}$, as well as some negative examples (i.e.\ individuals that are neither instances of $A$ nor $B$). To select these negative examples, we first sample instances from categories that have the same parent as $A$ and $B$, choosing as many such negative examples as we have positive examples. Second, we also sample the same number of negative examples from randomly selected categories in the taxonomy.

Let $F^1_{AB}$ be the F1 score achieved by the Gaussian classifier and $F^2_{AB}$ the F1 score of the GLR classifier. Our hypothesis is that $F^1_{AB} \ll F^2_{AB}$ suggests that $A$ and $B$ are conceptual neighbors, while $F^1_{AB} \gg F^2_{AB}$ suggests that they are not. This intuition is captured in the following score:
$$
s_{AB} = \frac{F^2_{AB}}{F^1_{AB}+F^2_{AB}}
$$
where we consider $A$ and $B$ to be conceptual neighbors if $s_{AB}\gg 0.5$.

%To find likely conceptual neighborhoods (that will be used to train the text classifier), we use the performance of the Gaussian and GLR classifiers described above as indicator. For each sibling pair $(S_i,S_j)$ in the training set, we output the F1-scores of the Gaussian classifier and the GLR classifier. The task we consider is, given an instance $s$, we want to decide whether this instance is correctly predicted to belongs to one of two  neighborhood concepts or none of them.  The negative examples that we consider are instances from other siblings of $S_i$ and $S_j$, instances from their parent concept, and instances from randomly generated concepts. 

%The F1-score relative to each classifier will be used as an indicator to find conceptual neighborhood. Intuitively, if the Gaussian classifier behaves better, this means that representing the concepts $S_i$ and $S_j$ as two ellipsoidal regions is enough to decide whether an instance belongs to $S_i$ or $S_j$ or none of them. Now if the GLP provides better F1-score, this means that $S_i$ and $S_j$ are likely to be conceptual neighborhoods, in the sense that, the this classifier is first able to distinguish between instances that belong to two concept conceptual neighborhood or not, and second to separate the the instances of each concept.  

%_____________________________
\subsection{Step 2: Predicting Conceptual Neighborhood from Text}\label{secTextPrediction}

We now consider the following problem: given two BabelNet categories $A$ and $B$, predict whether they are likely to be conceptual neighbors based on the sentences from a text corpus in which they are both mentioned. To train such a classifier, we use the distant supervision labels from Section \ref{secDistantSupervision} as training data. Once this classifier has been trained, we can then use it to predict conceptual neighborhood for categories for which only few instances are known.

To find sentences in which both $A$ and $B$ are mentioned, we rely on a disambiguated text corpus in which mentions of BabelNet categories are explicitly tagged. Such a disambiguated corpus can be automatically constructed, using methods such as the one proposed by \newcite{mancini-etal-2017-embedding}, for instance. For each pair of candidate categories, we thus retrieve all sentences where they co-occur. Next, we represent each extracted sentence as a vector. To this end, we considered two possible strategies:
%Given pre-trained word embeddings, e.g. Word2Vec \cite{DBLP:journals/corr/abs-1301-3781}, GloVe \cite{glove2014} or FastText \cite{FASTTEXT},
\begin{enumerate}
    \item \textbf{Word embedding averaging:} We compute a sentence embedding by simply averaging the word embeddings of each word within the sentence. Despite its simplicity, this approach has been shown to provide competitive results \cite{arora2016simple}, in line with more expensive and sophisticated methods e.g.\ based on LSTMs.

    \item \textbf{Contextualized word embeddings:} The recently proposed contextualized embeddings  \cite{peters-etal-2018-deep,devlin2018bert} have already proven successful in a wide range of NLP tasks. Instead of providing a single vector representation for all words irrespective of the context, contextualized embeddings predict a representation for each word occurrence which depends on its context. These representations are usually based on pre-trained language models. In our setting, we extract the contextualized embeddings for the two candidate categories within the sentence. To obtain this contextualized embedding, we used the last layer of the pre-trained language model, which has been shown to be most suitable for capturing semantic information \cite{peters2018dissecting,tenney2019bert}. We then use the concatenation of these two contextualized embeddings as the representation of the sentence.  
\end{enumerate}{}
For both strategies, we average their corresponding sentence-level representations across all sentences in which the same two candidate categories are mentioned. Finally, we train an SVM classifier on the resulting vectors to predict for the pair of siblings $(A,B)$ whether $s_{AB}> 0.5$ holds.

%_____________________________
\subsection{Step 3: Category Induction}\label{subsecUsingCN}
Let $C$ be a category and assume that $N_1,...,N_k$ are conceptual neighbors of this category. Then we can model $C$ by generalizing the idea underpinning the GLR classifier. In particular, we first learn a Gaussian distribution from all the instances of $C$ and $N_1,...,N_k$. This Gaussian model allows us to estimate the probability $P(C\cup N_1\cup ...\cup N_k \,|\, \mathbf{e})$ that $e$ belongs to one of $C,N_1,...,N_k$. If this probability is sufficiently high (i.e.\ higher than 0.5), we use a multinomial logistic regression classifier to decide which of these categories $e$ is most likely to belong to. Geometrically, we can think of the Gaussian model as capturing the relevant local domain, while the multinomial logistic regression model carves up this local domain, similar as in Figure \ref{figDiagram}.

In practice, we do not know with certainty which categories are conceptual neighbors of $C$. Instead, we select the $k$ categories (for some fixed constant $k$), among all the siblings of $C$, which are most likely to be conceptual neighbors, according to the text classifier from Section \ref{secTextPrediction}.

%We consider a set of concepts $C$ that have few instances, i.e less than 50 instances. For each target concept $T \in C$, we assume that a set of its siblings $S$ is available in $C$.  Let $\{S_1, ..., S_n\}$ be the set of siblings of the target concept $T$. Using text classifier results, we first rank the siblings of $T$ according to their confidence scores, and then select the top-k $\{N_1, ...,N_k\}$ most likely conceptual neighborhoods (\todo{otherwise we need to set up a threshold}). 
%Given an instance $s$, the task we consider is to decide whether $s$ belongs to the target concept $T$ knowing its conceptual neighborhoods. To this end, we first learn a Gaussian distribution on the instance of $T \cup N_1 \cup ... \cup N_k$. This distribution permits to model the region of $T \cup N_1 \cup ... \cup N_k$ and to estimate how likely an instance belongs the this region or not. To discriminate between instances of the target concept and its neighborhoods, we learn a multinomial logistic regression where each class model a concept in $T, N_1,...,N_k$. 

%******************************
\section{Experiments}
The central problem we consider is category induction: given some instances of a category, predict which other individuals are likely to be instances of that category. When enough instances are given, standard approaches such as the Gaussian classifier from Section \ref{secGaussian}, or even a simple SVM classifier, can perform well on this task. For many categories, however, we only have access to a few instances, either because the considered ontology is highly incomplete or because the considered category only has few actual instances. The main research question which we want to analyze is whether (predicted) conceptual neighborhood can help to obtain better category induction models in such cases. In Section \ref{secSetting}, we first provide more details about the experimental setting that we followed. Section \ref{secQuantitative} then discusses our main quantitative results. Finally, in Section \ref{secQualitative} we present a qualitative analysis.

\subsection{Experimental setting}\label{secSetting}
%\todo{Jose to explain text classifier}\\

\paragraph{Taxonomy} 

As explained in Section \ref{model}, we used BabelNet \cite{navigli2012babelnet} as our reference taxonomy. BabelNet is a large-scale full-fledged taxonomy consisting of heterogeneous sources such as WordNet \cite{Fellbaum:98}, Wikidata \cite{vrandevcic2014wikidata} and WiBi \cite{flati2016multiwibi}, making it suitable to test our hypothesis in a general setting.

\smallskip
\noindent \textbf{Vector space embeddings.} Both the distant labelling method from Section \ref{secDistantSupervision} and the category induction model itself need access to vector representations of the considered instances. To this end, we used the NASARI vectors\footnote{Downloaded from \url{http://lcl.uniroma1.it/nasari/}.}, which have been learned from Wikipedia and are already linked to BabelNet \cite{camacho2016nasari}.

\smallskip
\noindent \textbf{BabelNet category selection.} To test our proposed category induction model, we consider all BabelNet categories with fewer than 50 known instances. This is motivated by the view that conceptual neighborhood is mostly useful in cases where the number of known instances is small. For each of these categories, we split the set of known instances into 90\% for training and 10\% for testing. To tune the prior probability $\lambda_A$ for these categories, we hold out 10\% from the training set as a validation set. 

The conceptual neighbors among the considered test categories are predicted using the classifier from Section \ref{secTextPrediction}. To obtain the distant supervision labels needed to train that classifier, we consider all BabelNet categories with at least 50 instances. This ensures that the distant supervision labels are sufficiently accurate and that there is no overlap with the categories which are used for evaluating the model.

\smallskip
\noindent \textbf{Text classifier training.} As the text corpus to extract sentences for category pairs we used the English Wikipedia. In particular, we used the dump of November 2014, for which a disambiguated version is available online\footnote{Available at \url{http://lcl.uniroma1.it/sw2v}}. This disambiguated version was constructed using the shallow disambiguation algorithm of \newcite{mancini-etal-2017-embedding}. As explained in Section \ref{secTextPrediction}, for each pair of categories we extracted all the sentences where they co-occur, including a maximum window size of 10 tokens between their occurrences, and 10 tokens to the left and right of the first and second category within the sentence, respectively. For the averaging-based sentence representations we used the 300-dimensional pre-trained GloVe word embeddings \cite{glove2014}.\footnote{Pre-trained embeddings downloaded from \url{https://nlp.stanford.edu/projects/glove/}} 
To obtain the contextualized representations we used the pre-trained 768-dimensional BERT-base model \cite{devlin2018bert}.\footnote{We used the implementation available at \url{https://github.com/huggingface/pytorch-pretrained-BERT}}. 

The text classifier is trained on 3,552 categories which co-occur at least once in the same sentence in the Wikipedia corpus, using the corresponding scores $s_{AB}$ as the supervision signal (see Section \ref{secTextPrediction}). To inspect how well conceptual neighborhood can be predicted from text, we performed a 10-fold cross validation over the training data, removing for this experiment the \textit{unclear} cases (i.e., those category pairs with $s_{AB}$ scores between $0.4$ and $0.6$). We also considered a simple baselineWE based on the number of co-occurring sentences for each pairs, which we might expect to be a reasonably strong indicator of conceptual neighborhood, i.e.\ the more often two categories are mentiond in the same sentence, the more likely that they are conceptual neighbors. The results for this cross-validation experiment are summarized in Table \ref{crossvalidation}. Surprisingly, perhaps, the word vector averaging method seems more robust overall, while being considerably faster than the method using BERT. The results also confirm the intuition that the number of co-occurring sentences is positively correlated with conceptual neighborhood, although the results for this baseline are clearly weaker than those for the proposed classifiers.

%However, this is not the case for the category induction experiments, as we will show in the following sections. 
%Finally, despite the relatively low figures by all models, the performance of this model is competitive enough for our goal of accurately predicting high-confidence conceptual neighbors. 
%\red{To re-read and check}  

\begin{table}[]
\centering
\begin{tabular}{lrrrr} 
\toprule
\multicolumn{1}{c}{} & \multicolumn{1}{c}{\textbf{Acc}} & \multicolumn{1}{c}{\textbf{F1}} & \multicolumn{1}{c}{\textbf{Pr}} & \multicolumn{1}{c}{\textbf{Rec}} \\
\midrule
\textbf{Avg.}                 & \textbf{70.6}                        & \textbf{69.0}                         & \textbf{69.4}                       & \textbf{69.0}                     \\ 
\textbf{BERT}                 & 66.9                        & 65.8                         & 65.9                       & 66.2                     \\ 
\textbf{\#sents}          & 61.6                        & 46.6                         & 43.3                       & 54.3                   \\
\bottomrule
\end{tabular}

\caption{Cross-validation results on the training split of the text classifier (accuracy and macro-average F1, precision and recall).}
\label{crossvalidation}
\end{table}

%
%\paragraph{Pre-trained embeddings}
%

\smallskip
\noindent \textbf{Baselines.} To put the performance of our model in perspective, we consider three baseline methods for category induction. First, we consider the performance of the Gaussian classifier from Section \ref{secGaussian}, as a representative example of how well we can model each category when only considering their given instances; this model will be referred to as \textit{Gauss}. Second, we consider a variant of the proposed model in which we assume that all siblings of the category are conceptual neighbors; this model will be referred to as \textit{Multi}. Third, we consider a variant of our model in which the neighbors are selected based on similarity. To this end, we represent each BabelNet as their vector from the NASARI space. From the set of siblings of the target category $C$, we then select the $k$ categories whose vector representation is most similar to that of $C$, in terms of cosine similarity. This baseline will be referred to as \textit{Similarity}$_k$, with $k$ the number of selected neighbors. 

We refer to our model as \textit{SECOND-WEA}$_k$ or \textit{SECOND-BERT}$_k$ (SEmantic categories with COnceptual NeighborhooD), depending on whether the word embedding averaging strategy is used or the method using BERT.
%_____________________________
\subsection{Quantitative Results}\label{secQuantitative}

\begin{table}
\centering
\begin{tabular}{lllll}
\toprule
     &  & \textbf{Pr} & \textbf{Rec} & \textbf{F1} \\ 
\midrule     
\textit{Gauss}  &  & 23.0   &  27.4   &  22.3  \\ 
\textit{Multi}   &  & 37.7  & 75.2    &  44.2  \\ 
\midrule
\textit{Similarity\textsubscript{1}} &  & 28.7 & 69.2 &  33.8          \\ 
\textit{Similarity\textsubscript{2}} &  & 30.0  & 68.1   & 34.0   \\ 
\textit{Similarity\textsubscript{3}} &  &  31.6  &   67.2  &  34.3  \\ 
\textit{Similarity\textsubscript{4}} &  & 32.8 &  78.5  &  38.2     \\ 
\textit{Similarity\textsubscript{5}} &  &  37.2  & 80.6    & 42.8   \\ 

\midrule
\textit{SECOND-WEA\textsubscript{1}} &  & 32.7 & \textbf{90.1}   &  41.9      \\ 
\textit{SECOND-WEA\textsubscript{2}} &  & 42.2   &  82.6   &  49.3  \\ 
\textit{SECOND-WEA\textsubscript{3}} &  & 43.4   &   83.1  &  50.4  \\ 
\textit{SECOND-WEA\textsubscript{4}} &  & \textbf{47.7}   & 84.2    &  \textbf{54.2}  \\ 
\textit{SECOND-WEA\textsubscript{5}} &  & 44.0   &  82.6   & 51.1   \\ 
\midrule
\textit{SECOND-BERT\textsubscript{1}} & & 38.5  &  87.1  &   47.0      \\ 
\textit{SECOND-BERT\textsubscript{2}} &  & 43.9  &  84.1   & 50.8   \\ 
\textit{SECOND-BERT\textsubscript{3}} &  &  44.9  &   84.4  & 52.2   \\ 
\textit{SECOND-BERT\textsubscript{4}} &  & 46.2   & 85.4    & 53.3   \\ 
\textit{SECOND-BERT\textsubscript{5}} &  &  43.8  &   84.7  &  51.3  \\
\bottomrule
\end{tabular}
\caption{Results (\%) of the category induction experiments \label{tabResults}}
\end{table}

Our main results for the category induction task are summarized in Table \ref{tabResults}. In this table, we show results for different choices of the number of selected conceptual neighbors $k$, ranging from 1 to 5.
As can be seen from the table, our approach substantially outperforms all baselines, with \textit{Multi} being the most competitive baseline. Interestingly, for the \textit{Similarity} baseline, the higher the number of neighbors, the more the performance approaches that of \textit{Multi}. The relatively strong performance of \textit{Multi} shows that using the siblings of a category in the BabelNet taxonomy is in general useful. However, as our results show, better results can be obtained by focusing on the predicted conceptual neighbors only. It is interesting to see that even selecting a single conceptual neighbor is already sufficient to substantially outperform the Gaussian model, although the best results are obtained for $k=4$. Comparing the \textit{WEA} and \textit{BERT} variants, it is notable that \textit{BERT} is more successful at selecting the single best conceptual neighbor (reflected in an F1 score of 47.0 compared to 41.9). However, for $k \geq 2$, the results of the \textit{WEA} and \textit{BERT} are largely comparable.

\begin{figure}[t]
\centering
\includegraphics[width=\linewidth]{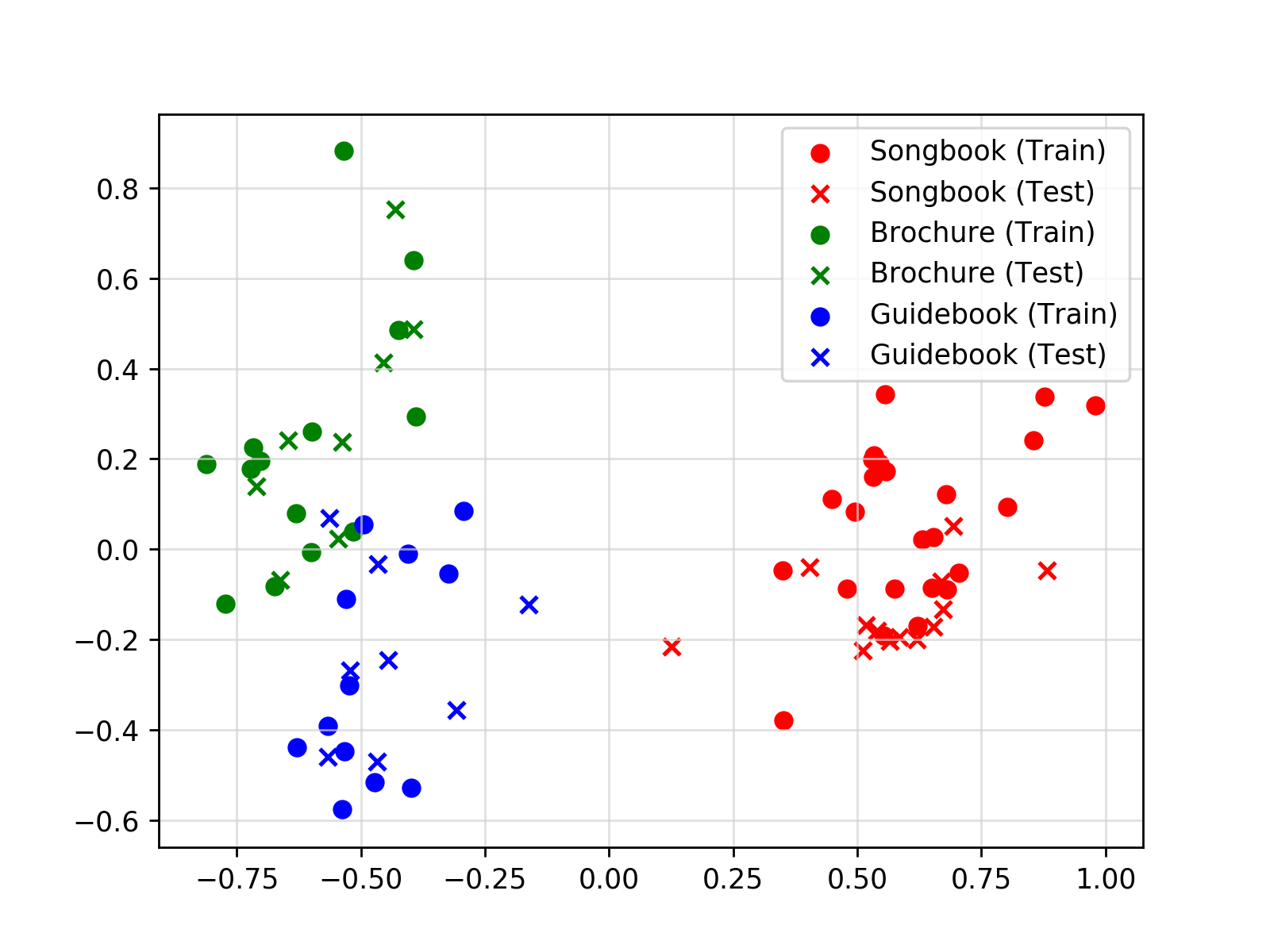}
\caption{Instances of three BabelNet categories which intuitively can be seen as conceptual neighbors.}\label{plot}
\end{figure}
%_____________________________
\subsection{Qualitative Analysis}\label{secQualitative}
To illustrate how conceptual neighborhood can improve classification results, Fig.\ \ref{plot} shows the two first principal components of the embeddings of the instances of three BabelNet categories: \emph{Songbook}, \emph{Brochure} and \emph{Guidebook}. All three categories can be considered to be conceptual neighbors. Brochure and Guidebook are closely related categories, and we may expect there to exist borderline cases between them. This can be clearly seen in the figure, where some instances are located almost exactly on the boundary between the two categories. On the other hand, Songbook is slightly more separated in the space. Let us now consider the left-most data point from the Songbook test set, which is essentially an outlier, being more similar to instances of Guidebook than typical Songbook instances. When using a Gaussian model, this data point would not be recognised as a plausible instance. When incorporating the fact that Brochure and Guidebook are conceptual neighbors of Songbook, however, it is more likely to be classified correctly.

To illustrate the notion of conceptual neighborhood itself, Table \ref{tabExamplesConceptualNeighbors} displays some selected category pairs from the training set (i.e.\ the category pairs that were used to train the text classifier), which intuitively correspond to conceptual neighbors. The left column contains some selected examples of category pairs with a high $s_{AB}$ score of at least 0.9. As these examples illustrate, we found that a high $s_{AB}$ score was indeed often predictive of conceptual neighborhood. As the right column of this table illustrates, there are several category pairs with a lower $s_{AB}$ score of around 0.5 which intuitively still seem to correspond to conceptual neighbors. When looking at category pairs with even lower scores, however, conceptual neighborhood becomes rare. Moreover, while there are several pairs with high scores which are not actually conceptual neighbors (e.g.\ the pair \textit{Actor} -- \textit{Makup Artist}), they tend to be categories which are still closely related. This means that the impact of incorrectly treating them as conceptual neighbors on the performance of our method is likely to be limited. On the other hand, when looking at category pairs with a very low confidence score we find many unrelated pairs, which we can expect to be more harmful when considered as conceptual neighbors, as the combined Gaussian will then cover a much larger part of the space. Some examples of such pairs include \textit{Primary school} -- \textit{Financial institution}, \textit{Movie theatre} -- \textit{Housing estate}, \textit{Corporate title} -- \textit{Pharaoh} and \textit{Fraternity} -- \textit{Headquarters}. 

\begin{table}[t]
\resizebox{\columnwidth}{!}{
%\footnotesize
\centering
\begin{tabular}{r@{\hspace{3pt}}c@{\hspace{3pt}}lr@{\hspace{3pt}}c@{\hspace{3pt}}l}
\toprule
\multicolumn{3}{c}{\textbf{High confidence}} & \multicolumn{3}{c}{\textbf{Medium confidence}}\\
\cmidrule(lr){1-3}  \cmidrule(lr){4-6}
Actor & -- & Comedian               &  Cruise ship &--& Ocean liner\\
Journal &--& Newspaper            &  Synagogue &--& Temple\\
Club &--& Company                 &  Mountain range &--& Ridge\\
Novel &--& Short story            &  Child &--& Man\\
Tutor &--& Professor              &  Monastery &--& Palace\\
Museum &--& Public aquarium       &  Fairy tale &--& Short story\\
Lake &--& River                   &  Guitarist &--& Harpsichordist\\
\bottomrule
\end{tabular}
}
\caption{Selected examples of siblings $A$--$B$ for which the conceptual neighborhood score $s_{AB}$ is higher than 0.9 (left column) and around 0.5 (right column).\label{tabExamplesConceptualNeighbors}}
\end{table}

%%%%

Finally, in Tables \ref{tabLowConceptualNeighbors} and \ref{tabHighConceptualNeighbors}, we show examples of the top conceptual neighbors that were selected for some categories from the test set. Table \ref{tabLowConceptualNeighbors} shows examples of BabelNet categories for which the F1 score of our SECOND-WEA$_1$ classifier was rather low. As can be seen, the conceptual neighbors that were chosen in these cases are not suitable. For instance, \emph{Bachelor's degree} is a near-synonym of \emph{Undergraduate degree}, hence assuming them to be conceptual neighbors would clearly be detrimental. In contrast, when looking at the examples in Table \ref{tabHighConceptualNeighbors}, where categories are shown with a higher F1 score, we find examples of conceptual neighbors that are intuitively much more meaningful. %This supports the overall conclusion that conceptual neighborhood is important and helpful for learning higher-quality concept representations.

\begin{table}[t]
\resizebox{\columnwidth}{!}{
%\footnotesize
\centering
\begin{tabular}{lll}
\toprule
\textbf{Concept} & \textbf{Top neighbor} & \textbf{F1}\\ 
\cmidrule(lr){1-3}  
Bachelor's degree  &  Undergraduate degree   & 34  \\ 
Episodic video game  &   Multiplayer gamer  &  34  \\ 
501(c) organization  &   Not-for-profit arts organization  & 29 \\ 
Heavy bomber  &  Triplane  & 41  \\ 
Ministry  &  United States government   &   33 \\
\bottomrule
\end{tabular}
}
\caption{Top conceptual neighbors selected for categories associated with a low F1 score.\label{tabLowConceptualNeighbors}}
\end{table}

\begin{table}[t]
\resizebox{\columnwidth}{!}{
%\footnotesize
\centering
\begin{tabular}{llll}
\toprule
\textbf{Concept} & \textbf{Top neighbor} & \textbf{F1} \\ 
\cmidrule(lr){1-3}  
Amphitheater ~~~~~~~~~  ~~~~~~~~~ &  Velodrome ~~~~~~~~~ ~~~~~~~~~   & 67 \\ 
Proxy server  &   Application server  &  61  \\ 
Ketch   &   Cutter  &  74  \\ 
Quintet   &  Brass band   & 67  \\ 
Sand dune  &  Drumlin   &   71 \\ 
\bottomrule
\end{tabular}
}
\caption{Top conceptual neighbors selected for categories associated with a high F1 score.\label{tabHighConceptualNeighbors}}
\end{table}

%******************************
\section{Conclusions}
We have studied the role of conceptual neighborhood for modelling categories, focusing especially on categories with a relatively small number of instances, for which standard modelling approaches are challenging. To this end, we have first introduced a method for predicting conceptual neighborhood from text, by taking advantage of BabelNet to implement a distant supervision strategy. We then used the resulting classifier to identify the most likely conceptual neighbors of a given target category, and empirically showed that incorporating these conceptual neighbors leads to a better performance in a category induction task.

In terms of future work, it would be interesting to look at other types of lexical relations that can be predicted from text. One possible strategy would be to predict conceptual betweenness, where a category $B$ is said to be between $A$ and $C$ if $B$ has all the properties that $A$ and $C$ have in common \cite{DBLP:conf/ijcai/SchockaertL18} (e.g.\ we can think of \emph{wine} as being conceptually between \emph{beer} and \emph{rum}). In particular, if $B$ is predicted to be conceptually between $A$ and $C$ then we would also expect the region modelling $B$ to be between the regions modelling $A$ and $C$.

\smallskip
\noindent\textbf{Acknowledgments.}
Jose Camacho-Collados, Luis Espinosa-Anke and Steven Schockaert were funded by ERC Starting Grant 637277. Zied Bouraoui was supported by CNRS PEPS INS2I MODERN.

\bibliography{commonsense,entity,wordembedding}
\bibliographystyle{aaai}
\end{document}